# AT-ViT: Area-Targeted Multi-View Vision Transformer with Cross-Attention and Multi-Scale Patching for Plant Trait Recognition in Herbarium Images

Amani Sedrat[a], Takieddine Chehhat[a], Youcef Sklab[b,1], Hanane Ariouat[c], Abderrazak Sebaa[a], Eric Chenin[b], Jean-Daniel Zucker[b], Edi Prifti[b]

*[a]Laboratoire LITAN, École supérieure en Sciences et Technologies de l'Informatique et du Numérique RN 75, Amizour 06300, Béjaïa, Algérie*
*[b]IRD, Sorbonne Université, UMMISCO, Paris, France*
*[c]Institut de Biomecanique Humaine Georges Charpak, Arts et Metiers Institute of Technology, 75013 Paris, France*

**Abstract**

Automated plant traits recognition from herbarium images is essential for plant sciences, yet remains challenging because background elements (e.g., textual labels, mounting artifacts, and color charts) can introduce shortcut learning, leading models to rely on spurious non-plant cues rather than plant morphology. This bias degrades both generalization and interpretability. In this paper, we introduce **AT-ViT**, a dual-branch Vision Transformer that jointly encodes raw herbarium scans and their segmented-derived counterparts via a multi-scale, multi-view cross-attention fusion scheme. AT-ViT further incorporates a mask-guided patch weighting mechanism that amplifies plant-relevant regions and attenuates background-driven features. By learning from the original scans while being guided by segmentation masks through the mask-guided patch reweighting mechanism, the model is encouraged to focus on plant organs and learn plant-centric representations more effectively. Across multiple trait classification tasks (e.g., leaf base shape, thorns), AT-ViT delivers consistent accuracy gains, improves attention localization on plant regions, and exhibits increased robustness under synthetic background perturbations. Specifically, AT-ViT substantially improves spatial attention grounding, boosting plant-region alignment (Avg IoU_p: +15.66 to +18.03 pp) while reducing background overlap (Avg IoU_b: −27.92 to −31.02 pp) relative to CrossViT, and remains markedly more robust to background perturbations, outperforming ResNet101 by up to +32.32 accuracy points and CrossViT by up to +5.07 points under background-noise conditions.

## 1. Introduction

Digitized herbarium specimens (Figure 1) have become an essential resource for plant science, enabling large-scale research in taxonomy, trait evolution, biodiversity monitoring, and phenological analysis. Initiatives such as GBIF[2] and ReColNat[3] have made millions of herbarium images accessible, providing visual records of plant diversity spanning centuries, continents, and taxa [13, 26]. However, despite their scientific value, herbarium scans pose distinct challenges for automated image analysis. Non-biological visual elements (such as specimen labels, barcodes, mounting artifacts, and color charts) introduce clutter that can mislead deep learning models and distort the representations they learn.

Deep learning, particularly Convolutional Neural Networks (CNNs) and Vision Transformers (ViTs), has enabled significant progress in herbarium image classification [27, 23, 2, 4, 3, 22, 20]. Yet, recent studies show that these models often focus on background regions rather than biologically relevant plant structures [15, 7, 2], undermining robustness, generalizability, and interpretability. This issue is especially critical for fine-grained traits, such as leaf base shape or the presence of small thorns, which require precise focus on localized plant organs.

[1] Corresponding author: Youcef Sklab - Email: youcef.sklab@ird.fr

[2] https://www.gbif.org

[3] https://recolnat.fr

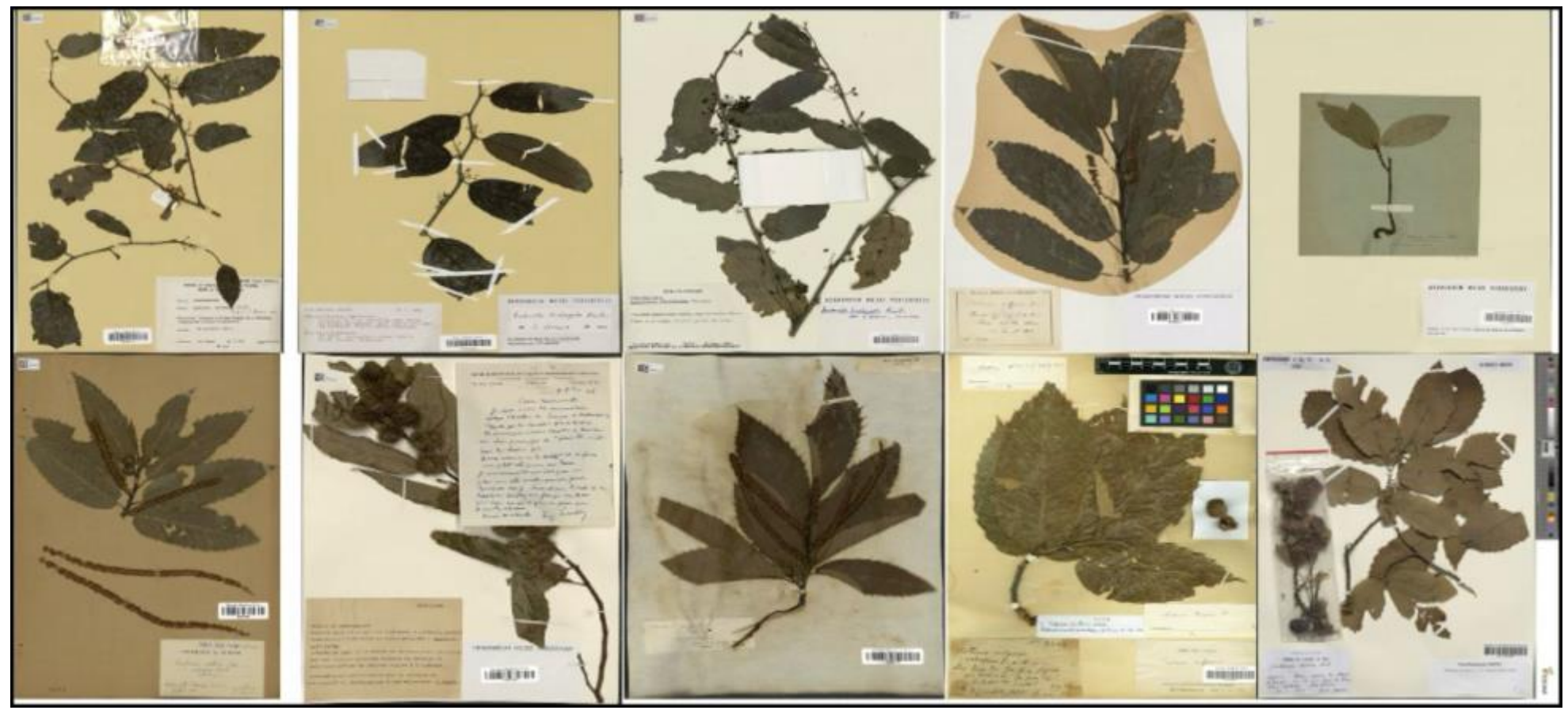

Figure 1: Representative herbarium scans, obtained from the ReColNat platform, illustrating the visual complexity of digitized specimens. Non-biological elements (such as labels, mounting tape, scale bars, and color charts) introduce background artifacts that can induce shortcut learning, biasing model attention toward suprious non-plant cues rather than plant morphology. In addition, dried plant material often occupies only a small portion of the total image area, making trait recognition especially challenging.

To address these limitations, segmentation-based approaches have gained traction. Methods like U-Net and PlantSAM [22] aim to isolate plant regions from the background, helping guide models toward meaningful structures. Notably, PlantSAM reports that, on average, only 20% of the image area typically contains plant material, explaining why models trained on unsegmented images tend to overfit to irrelevant background patterns. Despite segmentation's utility in isolating plant content, relying solely on segmented inputs can be limiting. First, segmentation masks are not always accurate: they may leave residual pixels from the background or introduce structured artifacts (e.g., black regions after background removal) that models can still exploit. Second, relying exclusively on segmented images can be limiting because segmentation may remove subtle but trait-relevant details such as fine venation, tiny prickles, trichomes, or petiole boundaries, and may distort spatial relationships that matter for trait interpretation. As illustrated in Figure 2, certain regions of the plant can even be omitted during the segmentation process, leading to the loss of valuable traits information.

Importantly, even with accurate segmentation, models may still exhibit background-driven localization biases, as shown in Figure 3. This suggests that networks can encode spurious correlations from residual low-frequency cues or dataset priors learned during training and may even learn systematic patterns introduced by preprocessing (e.g., consistent black backgrounds). These observations motivate a central question: how can we enforce learning on the plant while minimizing background reliance, without discarding the richer and more faithful visual information contained in the original scans? Maintaining access to the original scan remains essential for preserving faithful morphological cues. Other solutions, such as SIM-Net [20], incorporate segmented 2D views and inferred 3D point clouds to improve trait classification. However, their reliance on 3D reconstruction limits scalability in practical applications.

To overcome these challenges, we propose AT-ViT, a multi-view, multi-scale dual-branch Vision Transformer built upon the CrossViT architecture [5]. It is motivated by the need to preserve the rich morphological information contained in raw herbarium scans while explicitly constraining the model to attend to plant regions. To this end, we treat the original image as the primary source of visual evidence and leverage segmentation masks as an auxiliary spatial prior that approximately delineates the region of interest, even when boundaries are imperfect. Building upon CrossViT [5], AT-ViT jointly encodes the raw scan and its segmentation-derived view. The architecture incorporates a segmentation-guided patch weighting module that reweights token contributions to attenuate background-driven features and emphasize plant-relevant regions. A dedicated cross-attention fusion mechanism then integrates both branches, enabling the model to combine fine-grained texture cues and faithful appearance information from the original scan with the cleaner structural guidance provided by segmentation. We demonstrate that AT-ViT significantly improves:

- classification accuracy for fine-grained morphological traits,
- alignment of attention maps with plant regions (measured via IoU),
- and robustness to synthetic background noise, further indicating that AT-ViT learns plant-centric representations rather than relying on background-driven spurious cues.

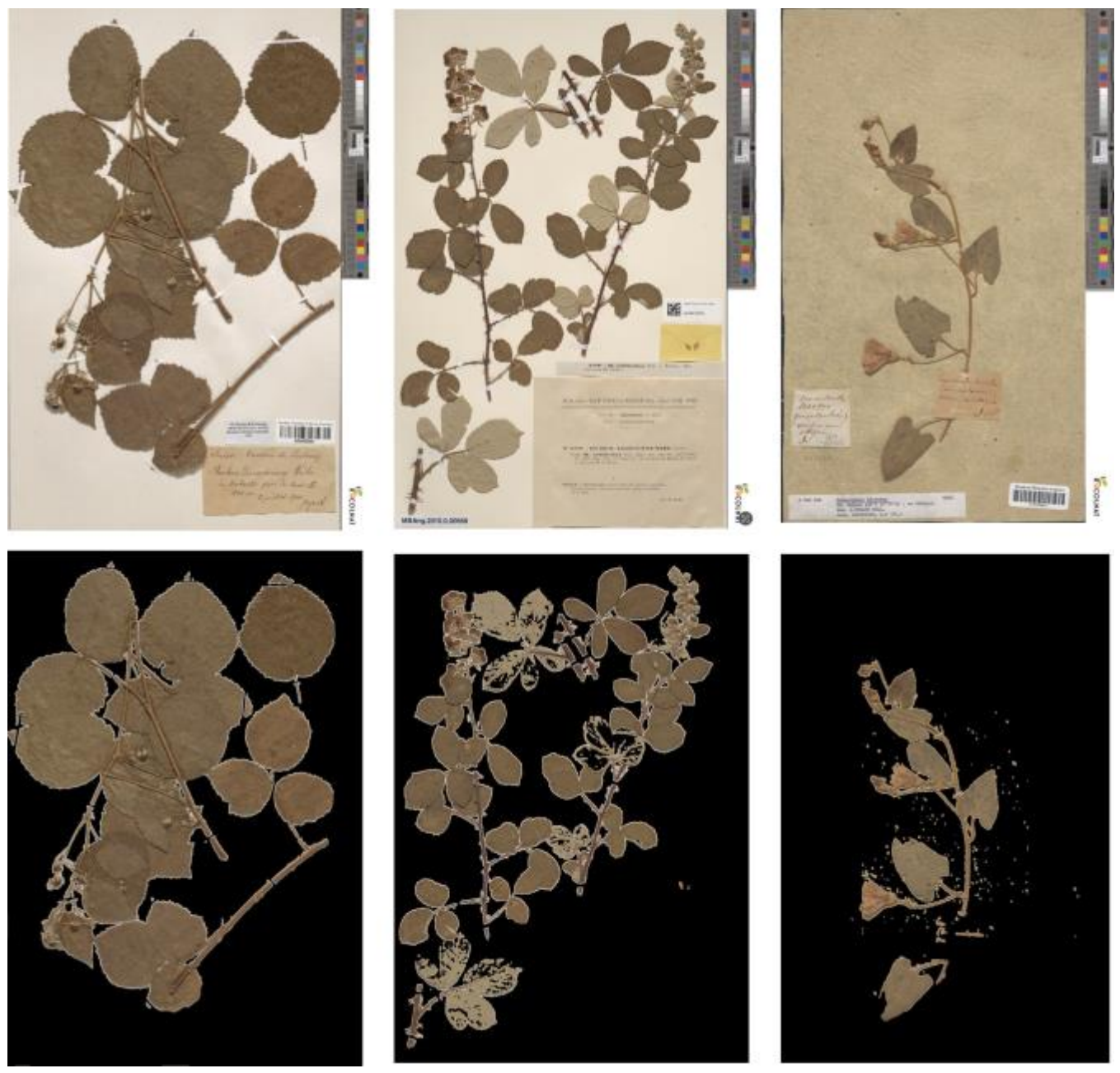

Figure 2: Illustrations of segmentation results for three herbarium specimens from ReColNat. The top row displays the original scans; the bottom row shows the corresponding segmented images. For specimen AIX055881 (left), the segmentation is highly accurate, cleanly isolating the plant from the background. ANG015003 (middle) is generally well segmented, with good contour preservation, though a few leaves are partially missed. In contrast, AV0009941 (right) exhibits segmentation errors, where parts of the background remain due to the mounting paper's texture and color being similar to that of the plant. These examples highlight both the strengths and limitations of the segmentation pipeline and support the use of both raw and segmented inputs in AT-ViT.

AT-ViT thus bridges the gap between segmentation-based preprocessing and transformer-based trait classification, offering a scalable and interpretable framework for image-based plant phenotyping. We hypothesize that explicitly guiding transformer attention through segmentation-informed patch weighting enhances both predictive performance and interpretability in fine-grained morphological trait recognition.

The remainder of this paper is organized as follows. Section 2 reviews related work on deep learning approaches for plant trait recognition, focusing on convolutional neural networks, vision transformers, and segmentation techniques in herbarium image analysis. Section 3 details the proposed AT-ViT architecture, including its dual-branch design, patch-weighting mechanism, and cross-attention strategy for integrating raw and segmented inputs. Section 4 describes the experimental setup, including the dataset, training protocols, and evaluation metrics used to assess the model's performance. Section 5 presents quantitative and qualitative results, comparing AT-ViT against the baseline CrossViT model and analyzing its robustness to synthetic noise. Finally, Section 6 summarizes the key findings and discusses future research directions for enhancing the model's interpretability and scalability.

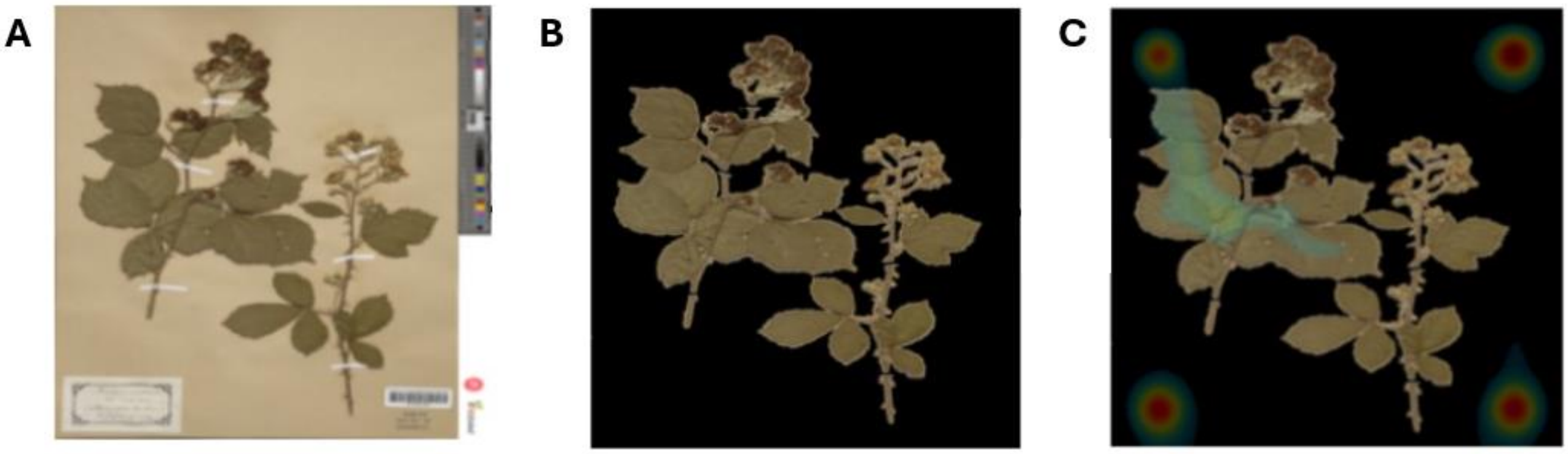

Figure 3: DL models can retain spurious localization biases even when the background is explicitly removed. From left to right: (1) the original herbarium scan (source: ReColNat); (2) the preprocessed image after binary segmentation; (3) the Grad-CAM heatmap (ResNet101) computed at inference time. Despite the suppression of background content, the Grad-Cam heatmap still shows substantial responses in peripheral and semantically irrelevant regions, indicating residual reliance on non-plant cues.

## 2. Related Work

The intersection of computer vision and plant science has seen significant progress with the advent of large-scale digitized herbarium datasets. These digital collections, enriched with centuries of botanical data, have enabled a wide range of applications, from taxonomic classification to phenological trait extraction [13, 26]. Central to this progress has been the adoption of deep learning models.

Early efforts predominantly relied on Convolutional Neural Networks (CNNs). Foundational studies such as [8] and [27] demonstrated that CNNs can effectively classify plant taxa and leaf-related traits directly from herbarium images. Subsequent work extended CNN-based pipelines beyond global classification to organ-level localization, detection, and segmentation (e.g., leaves and stems) across diverse sheet layouts and mounting conditions [23, 24]. These approaches established a practical basis for trait-oriented phenotyping by enabling region-specific inference and reducing the confounding effect of irrelevant areas. In parallel, object detection models—particularly YOLO variants—have been used to separate biological structures from non-biological elements [3]. For instance, [3] enhanced YOLOv7 with an attention-gated mechanism tailored to herbarium imagery and reported strong performance in organ detection under severe visual clutter, highlighting the importance of domain-aware architectural adaptations when labels, barcodes, and mounting artifacts are pervasive. Nevertheless, CNNs and standard detectors can be limited by their locality biases: the predominance of local receptive fields may hinder modeling of traits expressed through long-range spatial dependencies (e.g., branching patterns, compound leaves), motivating the use of architectures with more global context modeling.

Vision Transformers (ViTs) address this limitation by leveraging self-attention to capture long-range interactions. Prior work has benchmarked ViT-based models on plant-related datasets, reporting improvements in generalization and class balance relative to CNN baselines [17]. Similarly, [10] showed that ViTs can achieve competitive performance on fine-grained leaf classification, even under limited training data, suggesting enhanced capacity for capturing subtle morphological cues. However, several studies caution that ViTs can be particularly sensitive to background content in herbarium settings. Specifically, pretrained ViTs may allocate a substantial fraction of their attention to non-plant regions (e.g., mounting sheets, labels) unless explicit constraints or spatial priors are introduced [7, 15]. This behavior can degrade robustness and diminish interpretability (two properties that are crucial for scientific use-cases requiring plant-centered evidence). More broadly, such failures are consistent with shortcut learning, wherein models exploit spurious but predictive correlations (e.g., acquisition artifacts or collection-specific layouts) rather than learning the intended semantics [12]. Recent analyses indicate that masking, suppressing, or reweighting irrelevant regions can improve generalization in fine-grained recognition tasks [1], reinforcing the need for explicit spatial guidance in herbarium-based trait learning.

Segmentation has therefore emerged as a key component in herbarium image pipelines. Classical foreground–background separation methods (e.g., thresholding such as Otsu's method [16] and color-space filtering) have been used for coarse plant extraction. More recently, deep segmentation architectures (including U-Net [18] and DeepLabv3+ [6]) have enabled more accurate isolation of plant regions. For example, [9] employed U-Net++ for leaf segmentation and counting, while [24] reported strong segmentation performance for ferns using a ResNet34-based U-Net, achieving a Dice score of 96%. Ariouat et al. [4] proposed a comprehensive segmentation pipeline combining YOLO-based removal of non-biological objects, HSV filtering, morphological operations, and patch-wise U-Net segmentation with a ResNet101 backbone. Their results suggest that segmentation can improve downstream classification metrics and reduce background-driven saliency, thereby promoting plant-focused evidence. Building on detection-driven segmentation, Sklab et al. [22] introduced PlantSAM, which uses object detection cues to guide mask generation for herbarium specimens and improves the reliability of plant-region delineation for downstream recognition tasks. Importantly, [2] compared training on raw versus segmented herbarium images and showed that, in the absence of segmentation, even high-performing classifiers may rely on background cues; attribution analyses (e.g., Grad-CAM) indicated that segmentation can redirect model responses toward biologically meaningful structures, strengthening interpretability and scientific validity.

Complementary to segmentation-only pipelines, recent work has explored multi-modal and multi-branch architectures. Sklab et al. [20] proposed SIM-Net, which fuses 2D CNN features with 3D point-cloud representations inferred from segmented plant regions, reporting gains in accuracy and F1-score and suggesting that shape priors can improve robustness when paper texture resembles plant tissue. In another direction, [11] introduced ROI-ViT, a dual-branch transformer that integrates an image stream with a region-of-interest (ROI) stream via cross-attention. Although developed for pest classification and based on ROI boxes rather than pixel-level masks, this line of work supports the broader principle that spatial priors and guided fusion can mitigate background distraction.

Collectively, these studies motivate segmentation-guided transformer architectures that preserve the rich visual detail of raw scans while explicitly biasing the model toward plant regions. Our work builds on this literature by integrating segmentation-informed spatial priors into a multi-view transformer framework for fine-grained trait recognition in herbarium images.

## 3. Approach

In response to the challenges of morphological trait classification and the complexity of herbarium scans, we propose **AT-ViT**—a dual-branch transformer architecture inspired by CrossViT [5], but redesigned to support multi-view processing of herbarium images (Figure 4). Unlike the original CrossViT, which processes a single input, AT-ViT simultaneously processes two synchronized image views: the raw herbarium image and its segmented counterpart. This dual-input configuration allows the model to extract complementary visual cues. The raw image preserves fine-grained textures and biological context, while the segmented image emphasizes clean plant contours and shape abstractions.

Each view is processed by a dedicated transformer branch, forming a dual-branch architecture that allows feature extraction at different semantic granularities. The branches operate at different spatial resolutions, implementing a multi-scale strategy where image patches vary in size and embedding dimension. This design enhances the model's ability to capture both localized detail and global structure, which is critical for nuanced trait recognition in visually cluttered herbarium sheets.

- **Raw Image Branch (Small Branch)**: Receives a 240 × 240 RGB image, divides it into 12 × 12 patches (400 tokens), each embedded into 384-dimensional vectors.
- **Segmented Image Branch (Large Branch)**: Processes a 224 × 224 segmented image, split into 16 × 16 patches (196 tokens) with 768-dimensional embeddings.

This design choice is intentional. Assigning the *large branch* to the segmented input enables broader, semantic representation of plant morphology, leveraging the larger embedding dimension to encode high-level abstractions like overall leaf shape or branching structure. In contrast, the *small branch* is better suited to the raw image, as its finer-grained resolution allows it to capture local textures and resolve fine visual details, especially in cluttered regions where plant organs may partially overlap with labels, scale bars, or mounting artifacts. Maintaining this multiscale architecture (one of CrossViT's core strengths) ensures that the model benefits from both global context and localized precision. To improve interpretability and guide attention toward relevant regions, we introduce a segmentation-driven **patch weighting strategy**, relying on precomputed HSV-filtered U-Net masks from [4], we calculate per-patch saliency by measuring plant pixel density. Each 16 × 16 patch (from the 196-token grid) receives a *scalar* weight $w \in \mathbb{R}$ computed as:

$$w = \sigma(6 \cdot \text{ ratio } - 3) + \epsilon$$

Where $ratio \in [0,1]$ is the proportion of non-black pixels within the patch, derived from the binary plant segmentation mask, and $\sigma(\cdot)$ is the element-wise sigmoid activation, here used to non-linearly enhance contrast between densely and sparsely covered patches. The scaling $(6 \cdot ratio - 3)$ recenters the sigmoid curve to maximize discriminability in the mid-density range. During development, we experimented with several transformations to maximize discriminability between patch saliencies, and found that this specific scaling offered the best performance trade-off across classes. $\epsilon = 0.05$ is a small constant added to every weight for numerical stability, avoiding vanishing gradients or division-by-zero errors in subsequent normalization.

Let $\boldsymbol{E_p} \in \mathbb{R}^d$ denote the embedding vector of patch $p$ produced by the transformer patch embedding layer, where $d$ is the embedding dimension of the branch ($d$ = 768 for the segmented branch, $d$ = 384 for the raw branch). The weighted embedding $\mathbf{E}'$ is then obtained by broadcasting the scalar $w_p$ over all $d$ channels of the embedding:

$$\mathbf{E}_p' = w_p \cdot \mathbf{E}_p, \qquad p = 1,2,\dots,\mathrm{N}$$

where $N$ is the number of patches in the branch ($N$ = 196 for the segmented branch, $N$ = 400 for the raw branch after upsampling). To situate the patch weighting within the transformer's attention mechanism, we adopt the standard Query-Key-Value (QKV) formulation. The weighted embeddings $E_p'$ serve as input to each transformer encoder block. For self-attention within a branch, the standard multi-head attention operates as:

$$\text{Attention}(Q, K, V) = \text{softmax}\left(\frac{QK^T}{\sqrt{d_k}}\right)$$

where $Q = E'W_Q, K = E'W_K, V = E'W_V$, and $W_Q, W_K, W_V$ are learned projection matrices.

In practice:

- For the segmented image branch, each $w_p$ is multiplied directly with its corresponding 768-dimensional token embedding, producing a weighted feature matrix of size 196 × 768.

- For the raw image branch, the set of 196 scalar weights is upsampled via bilinear interpolation to match the 20 × 20 grid of the raw input (400 tokens), and each interpolated scalar is broadcast-multiplied with the corresponding 384-dimensional embedding.

The patch weights are computed once and are applied immediately after the linear patch embedding layer and before the first transformer encoder block in both branches (c.f. Figure 4). This weighting occurs only once per forward pass and is not reapplied in subsequent layers but propagates through all encoder blocks via the modified embeddings. The [CLS] token remains unweighted to preserve its role as a global aggregator. This modulation acts as a soft attention mechanism that selectively amplifies morphology-rich patches while suppressing background and irrelevant regions.

Each branch uses a transformer encoder, after which cross-attention is applied. For cross-attention between branches, the [CLS] token from one branch serves as the Query, while patch tokens from the other branch act as Keys and Values:

$$\mathrm{CrossAttn}_{\mathrm{small}\rightarrow\mathrm{large}} = \mathrm{softmax}\left(\frac{Q_{\mathrm{small}} K_{\mathrm{large}}^{T}}{\sqrt{d_k}}\right) V_{\mathrm{large}}$$

Where $Q_{\mathrm{small}} = [\mathrm{CLS}]_{\mathrm{small}} W_Q, \quad K_{\mathrm{large}} = E'_{\mathrm{large}} W_K, \quad V_{\mathrm{large}} = E'_{\mathrm{large}} W_V$ .

Symmetrically, the large branch [CLS] token queries the small branch patch tokens. This bidirectional cross-attention enables the model to align fine-grained textures from the raw image with clean morphological structures from the segmented image. The final [CLS] tokens are concatenated into a 1152-dimensional vector and passed to a fully connected head for binary classification.

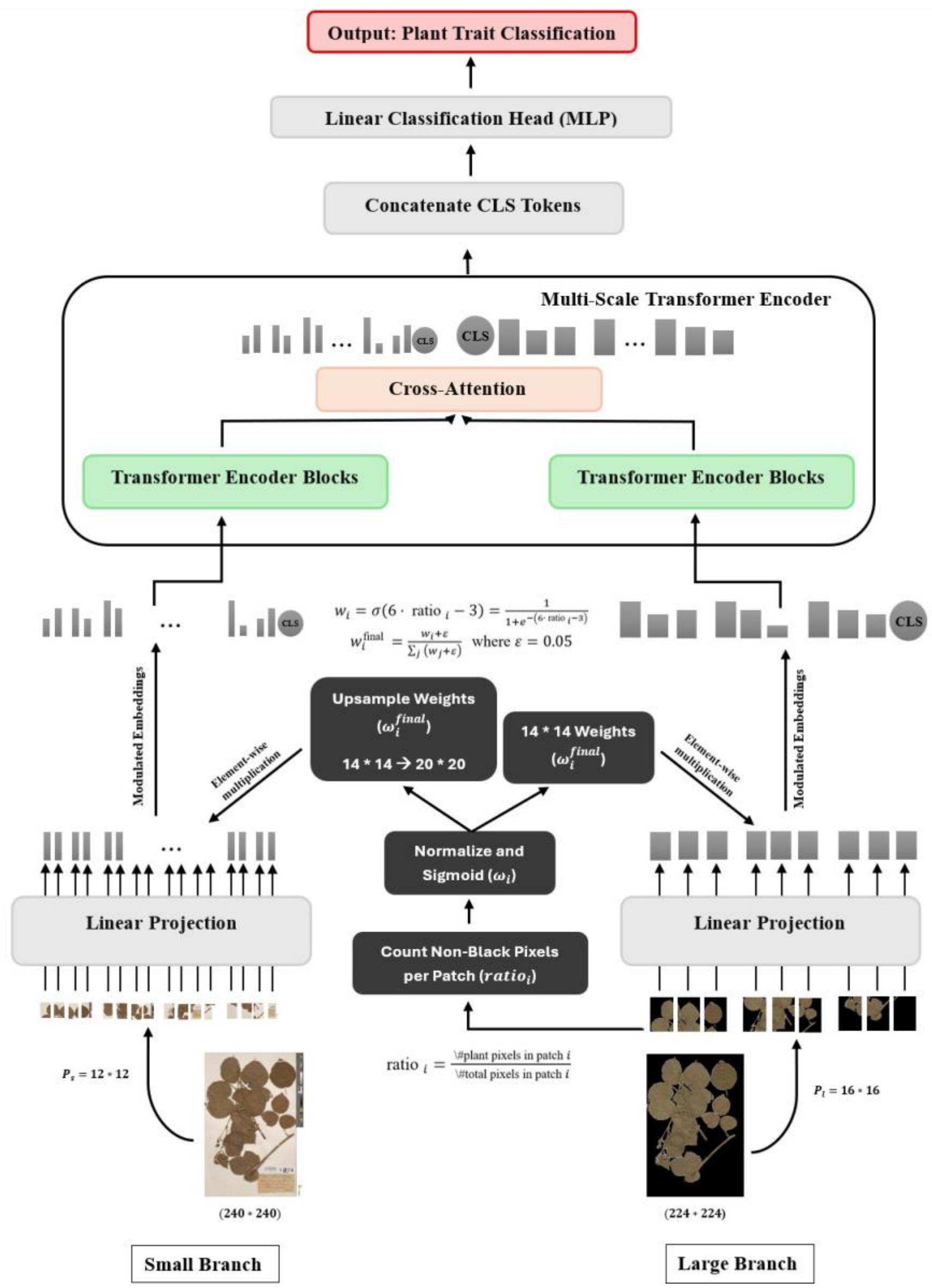

Figure 4: Schematic overview of the proposed AT-ViT architecture. The model processes raw herbarium images and their corresponding segmented versions through two parallel branches—referred to as the small and large branches, respectively. Each branch generates patch embeddings via linear projection, followed by Transformer encoder blocks. The segmented branch incorporates a patch-weighting mechanism, where weights are computed based on the proportion of non-black (plant) pixels per patch and modulated via a normalized sigmoid function. These weights are used to refine attention during token encoding. Outputs from both branches are fused in a multi-scale transformer encoder via cross-attention, and the concatenated CLS tokens are passed to a linear classification head for final trait prediction.

## 4. Experiments

### *4.1. Dataset*

The dataset we used in this study is derived from the ReColNat digitization platform [21], which provides high-resolution scans of herbarium specimens curated at the Muséum national d'Histoire naturelle (MNHN) in Paris. Each image is annotated with morphological trait information relevant to trait-based classification. We focus on three binary classification tasks, targeting the following traits: (1) presence of thorns, (2) leaves with an acute base, and (3) leaves with acuminate tips. For each specimen, a precomputed segmentation mask is also provided. These masks were generated using a U-Net-based segmentation pipeline developed by Ariouat et al. [4], designed to isolate plant material from background artifacts such as labels, mounting paper, and color scales. An overview of the number of annotated images per trait and their division into training and test sets is provided in Table 1.

These three morphological traits were selected based on data availability within our current annotated collection from ReColNat. While there is no specific ecological or taxonomic motivation for this particular combination, these traits represent different levels of visual complexity: thorns are relatively distinctive features (high inter-class separability), whereas acute leaf base and acuminate leaf tips require fine-grained discrimination of subtle shape variations at specific leaf regions. This diversity allows us to evaluate model performance across varying recognition difficulties. The primary motivation is to demonstrate that AT-ViT can improve spatial attention alignment across traits of different granularities.

Table 1: Annotated herbarium specimens used for each morphological trait. The dataset is divided into training and test sets for three binary classification tasks: presence of thorns, acute leaf base, and acuminate leaf tip.

| **Trait** | **Training Set** | **Test Set** |
|---|---|---|
| Presence of thorns | 1979 | 495 |
| Acute leaf base | 1797 | 450 |
| Acuminate leaf tip | 1812 | 453 |

### 4.2. *Training Details*

All models were trained using the (AdamW) optimizer [14] with a weight decay of 0.1 and a learning rate of 0.00005. Training was conducted for 70 epochs with a batch size of 16. To improve model generalization, standard data augmentation techniques were applied, including random horizontal flips, affine transformations, and color jittering. Image preprocessing was branch-specific: raw inputs were resized to 240×240 and segmented masks to 224×224, matching the resolution expectations of each branch in AT-ViT. All experiments were performed on a high-performance computing node equipped with 2×NVIDIA H100 GPUs (96 GB each), 192 CPU cores, and 768 GB of RAM. A step-based learning rate scheduler was used, decaying the rate by a factor of 0.5 every 5 epochs.

### 4.3. *Evaluation Metrics*

Model performance was assessed using classification accuracy, and to evaluate the attention alignment, we computed two Intersection-over-Union (IoU) metrics:

- **$\mathrm{IoU}_p$**: IoU between the binary attention mask and the ground truth *plant* mask.
- **$\mathrm{IoU}_b$**: IoU between the binary attention mask and the ground truth *background* mask.

Binary attention masks were obtained by thresholding the model's attention map at 0.5 (pixels with values $\geq 0.5$ were set to 1, others to 0).

$$\mathrm{IoU}_x = \frac{|A \cap G_x|}{|A \cup G_x|}, x \in \{p, b\} \tag{1}$$

Where:

- $A$: Binary attention mask.
- $G_p$: Ground truth plant mask (pixels labeled as plant areas = 1, others = 0).
- $G_b$: Ground truth background mask (pixels labeled as background areas = 1, others = 0).
- $|A \cap G_x|$: Number of pixels where both $A$ and $G_x$ are 1 (intersection).
- $|A \cup G_x|$: Number of pixels where at least one of $A$ or $G_x$ is 1 (union).

The **IoU** metrics (ranging from 0 to 1) directly quantify how the patch weighting mechanism influences spatial attention. By down-weighting background patches and amplifying plant-dense patches, the weighted embeddings modulate the attention computation such that attention scores are redistributed toward high-saliency regions after softmax normalization. This redistribution manifests as higher **IoU**$_p$ (stronger overlap with plant masks) and lower **IoU**$_b$ (reduced overlap with background masks), indicating that the transformer focuses on biologically relevant structures rather than spurious background patterns.

## 5. Results and Analysis

### 5.1. *Quantitative and Qualitative Evaluation of AT-ViT*

To evaluate the effectiveness of our proposed approach, we benchmark AT-ViT against two baseline architectures: ResNet101 and the standard CrossViT-Base [5] (crossvit_base_240). ResNet101 architecture provides a baseline representing traditional convolutional approaches, while CrossViT serves as our primary comparison as a vision transformer trained on raw herbarium images (224×224), without access to segmentation masks or patch-level weighting. Beyond classification accuracy, we assess how well each model aligns its attention with biologically meaningful plant regions using the Intersection over Union (IoU) metric, as described in the previous section. For a fair comparison, Grad-CAM [20] is used to generate heatmap visualizations for the ResNet model, while attention maps are extracted from the final transformer layer of the small branch in both AT-ViT and CrossViT.

Results summarized in Tables 2 and 3 reveal that AT-ViT achieves markedly superior spatial alignment across all three morphological traits compared to both baselines. The comparison with ResNet101 is particularly striking. As shown in Table 2, AT-ViT demonstrates improvements in plant region attention, with IoU$_p$ increasing by over 30 percentage points compared to ResNet101. While the ResNet101 model exhibits very low background attention (IoU$_b$ values between 10.18% and 14.10%), this does not translate to strong plant-focused attention (its IoU$_p$ values remain below 6.5% across all traits). This suggests that ResNet101 distributes its attention diffusely across the entire image rather than concentrating on either plant or background regions specifically. In contrast, AT-ViT achieves IoU$_p$ values above 37% while maintaining moderate background attention, indicating a more focused attention mechanism on morphologically relevant structures. When compared to CrossViT (Table 3), the average IoU with plant regions (Avg IoU$_p$) improves by more than 15 percentage points for each task, while the average IoU with background regions (Avg IoU$_b$) decreases by 27 to 31 points. Although CrossViT shows much higher background attention than ResNet101, it achieves better plant region alignment (IoU$_p$ around 20-21%) compared to ResNet101's diffuse attention patterns. Nevertheless, AT-ViT substantially outperforms CrossViT in directing attention toward morphologically relevant plant structures while simultaneously reducing background distractions.

Classification accuracy gains remain modest across both comparisons. Against ResNet101, accuracy improvements range from 0.21% to 8.17%. Compared to CrossViT, improvements range from +0.88% for "acuminate tips" to slight decreases for "acute base" and "thorns." However, the consistent and substantial improvement in spatial attention alignment across both baselines suggests that AT-ViT relies less on background artifacts and more on semantically grounded plant features. These findings support our hypothesis that segmentation-guided attention leads to more interpretable and robust model behavior. In this context, performance improvements are not merely numerical but reflect a deeper reorientation of the model's focus toward morphologically relevant regions—an essential step for building trustworthy AI in botanical image analysis.

Table 2: Comparison of classification accuracy, average IoU with the plant mask (Avg $IoU_p$), and average IoU with the background mask (Avg $IoU_b$) between the ResNet101 baseline and the proposed AT-ViT across three traits. Δ indicates the absolute change in performance (AT-ViT − ResNet101). AT-ViT demonstrates dramatic improvements in spatial attention alignment, with $IoU_p$ increasing by over 30 percentage points while maintaining moderate background attention.

| **Trait** | **ResNet101** | | | **AT-ViT** | | | **Δ** | | |
|---|---|---|---|---|---|---|---|---|---|
| | *Accuracy* | *Avg IoU$_p$* | *Avg IoU$_b$* | *Accuracy* | *Avg IoU$_p$* | *Avg IoU$_b$* | *Accuracy* | *Avg IoU$_p$* | *Avg IoU$_b$* |
| Leaves with acuminate tips | 69.75 | 5.30 | 10.52 | 77.92 | 39.04 | 43.32 | +8.17 | +33.74 | −32.80 |
| Leaves with an acute base | 77.55 | 5.76 | 10.18 | 79.33 | 37.00 | 46.91 | +1.78 | +31.24 | −36.73 |
| Thorns | 95.55 | 6.20 | 14.10 | 95.76 | 37.04 | 47.32 | +0.21 | +30.84 | −33.22 |

Table 3: Comparison of classification accuracy, average IoU with the plant mask (Avg $IoU_p$), and average IoU with the background mask (Avg $IoU_b$) between the CrossViT model and the proposed AT-ViT across three traits. Δ indicates the absolute change in performance (AT-ViT − CrossViT). While accuracy differences remain moderate, AT-ViT achieves a substantial improvement in spatial attention alignment by increasing $IoU_p$ and decreasing $IoU_b$.

| **Trait** | **CrossViT** | | | **AT-ViT** | | | **Δ** | | |
|---|---|---|---|---|---|---|---|---|---|
| | *Accuracy* | *Avg IoU$_p$* | *Avg IoU$_b$* | *Accuracy* | *Avg IoU$_p$* | *Avg IoU$_b$* | *Accuracy* | *Avg IoU$_p$* | *Avg IoU$_b$* |
| Leaves with acuminate tips | 77.04 | 21.01 | 74.34 | 77.92 | 39.04 | 43.32 | +0.88 | +18.03 | −31.02 |
| Leaves with an acute base | 79.77 | 20.88 | 74.93 | 79.33 | 37.00 | 46.91 | −0.44 | +16.12 | −28.02 |
| Thorns | 96.36 | 21.38 | 75.24 | 95.76 | 37.04 | 47.32 | −0.60 | +15.66 | −27.92 |

Figure 5 shows a visual comparison of the Grad-CAM of ResNet101 and the attention maps produced by CrossViT, and AT-ViT on a sample specimen for the trait *Leaves with acuminate tips*. The ResNet101 model demonstrates a diffuse attention patterns, achieving minimal plant region alignment ($IoU_p$ = 1.72%, $IoU_b$ = 20.90%). The CrossViT model exhibits high activation over background regions, reflected by a low $IoU_p$ (7.34%) and a very high $IoU_b$ (91.89%), indicating that its attention is largely misaligned with plant structures, though it performs better than the ResNet101 baseline in terms of plant region focus. In contrast, AT-ViT achieves better attention alignment, with $IoU_p$ reaching 46.43% and $IoU_b$ dropping to just 0.46%, demonstrating the effectiveness of segmentation-guided attention over both convolutional and transformer baseline architectures.

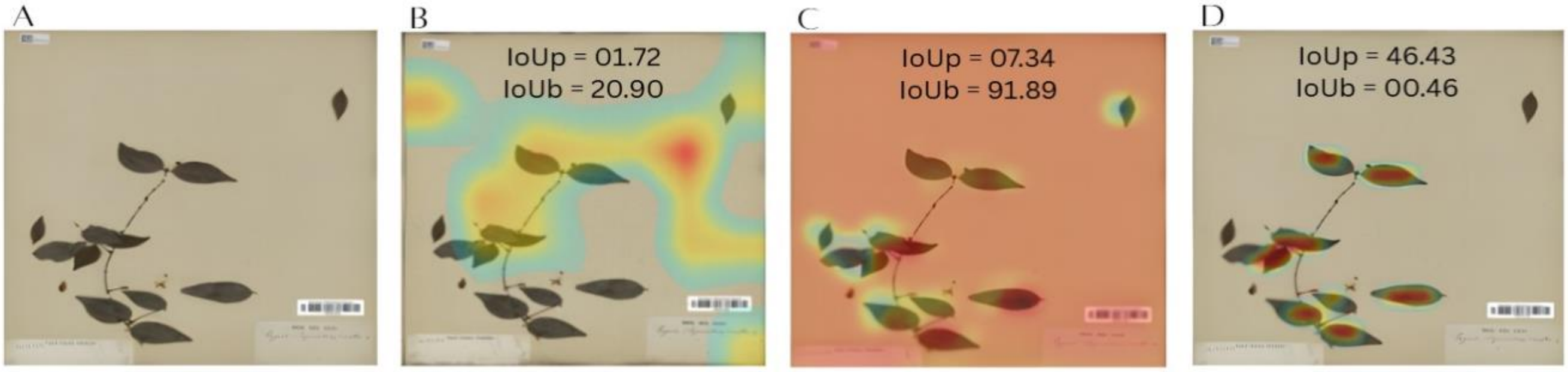

Figure 5: Illustration of areas of focus using the IoU metric for the trait *Leaves with acuminate tips*. (A) the original herbarium scan from ReColNat, (B) the Grad-CAM of ResNet101, (C) the attention map overlaid on the CrossViT model's output, and(D) the attention map from our AT-ViT model.

Figure 6 presents an example of a feature map, highlighting the regions considered salient by the model during the early stages of visual processing. In the Baselines, the focus is often diffuse and extends into non-informative background areas, such as labels and scale bars. In contrast, the patch-weighting strategy employed in AT-ViT results in more focused activations concentrated on the plant structures, effectively suppressing background artifacts and enhancing the model's sensitivity to biologically relevant regions.

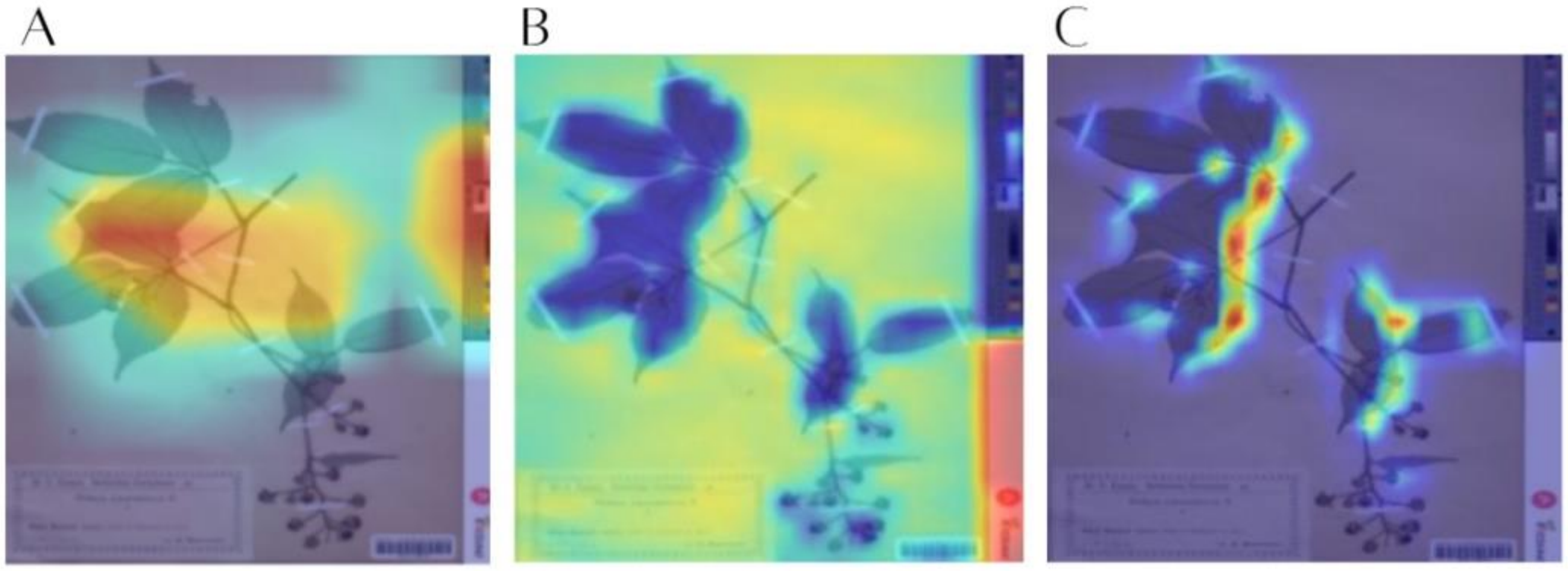

Figure 6: Feature maps computed over patch embeddings for CrossViT and AT-ViT and over the embeddings of the first convolutional layer of ResNet101 for the trait "thorns". **(A):** ResNet101 focuses on both plant and background. **(B)**: CrossViT showing dispersed activation over background. **(C)**: AT-ViT focuses strongly on plant regions due to saliency-aware reweighting.

Additionally, we also performed t-SNE visualizations of model embeddings for the thorns trait, illustrating the comparison between the baselines methods and our approach, with enhanced clustering and separation of classes Figure 7.

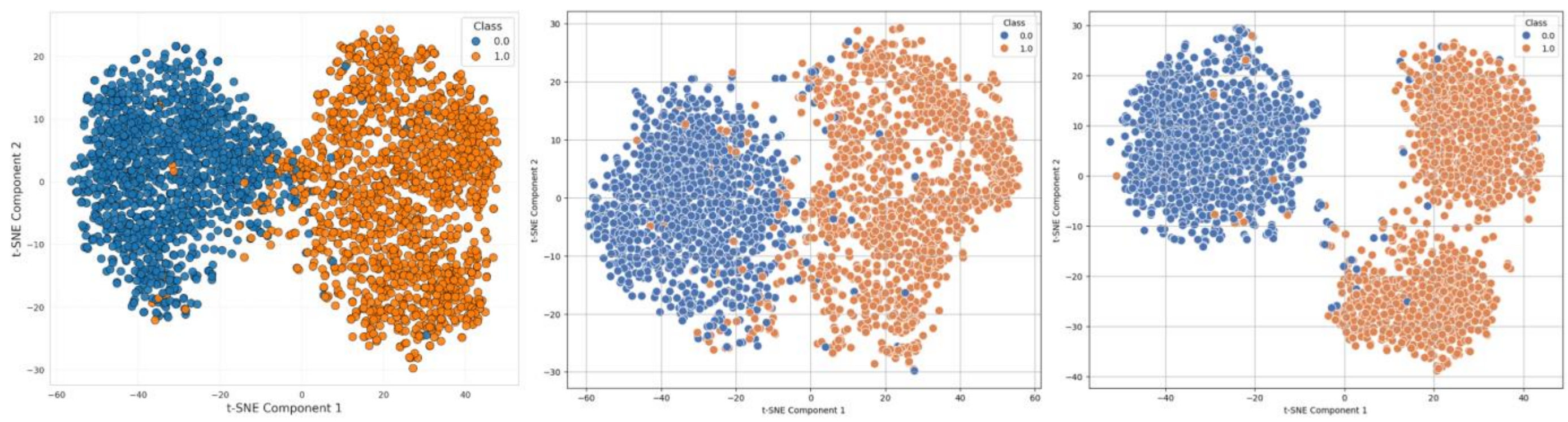

Figure 7: t-SNE visualization of learned feature embeddings for the trait *Presence of thorns*. Left: embeddings from the ResNet101 model. Middle: embeddings from the CrossViT model. Right: embeddings from our proposed AT-ViT. Each point represents an image in the dataset, colored by class label (blue = no thorns, orange = thorns). AT-ViT exhibits clearer class separation and denser intra-class clustering, indicating more discriminative feature representations.

Figure 8 illustrates the benefits of AT-ViT in enhancing attention localization, based on a comparative analysis of attention maps from the final transformer layers in CrossViT and AT-ViT, and Grad-CAM visualizations in ResNet101 for the trait "leaves with an acute base". In panel (A), the ResNet101 model demonstrates diffuse patterns with minimal localization to plant structures. In panel (B), the CrossViT model exhibits diffuse and poorly localized attention, frequently focusing on non-informative background regions such as mounting tape or scale bars—an artifact that often leads to incorrect predictions. Conversely, panel (C) shows that the AT-ViT model consistently allocates attention to morphologically meaningful plant structures, particularly around the base of the leaf, resulting in more accurate and interpretable outputs.

To further quantify these attention patterns across the dataset, we computed the average attention distribution over all test samples, as shown in Figure 9. This analysis reveals that ResNet101 emphasizes only a small area within the center of the image, CrossViT tends to emphasize peripheral and background areas, while AT-ViT concentrates attention near the center of the image. The difference in spatial resolution between the heatmaps arises from the nature of the underlying architectures and visualization methods. Grad-CAM applied to CNNs produces pixel-aligned localization maps derived from convolutional feature maps, whereas attention maps in Vision Transformers reflect interactions between discrete patch tokens, inherently limiting spatial resolution to the patch grid.

To assess whether this spatial focus corresponds to the actual location of plant material, we generated a pixel-wise density heatmap (Figure 10) by aggregating the binary segmentation masks across all samples. The resulting visualization confirms that plant pixels are predominantly concentrated in a big central region of the images, supporting the relevance and effectiveness of the spatial alignment exhibited by AT-ViT.

### 5.2. Robustness Evaluation: Assessing Model Dependence on Plant vs. Background Regions

To assess whether our model relies more on plant structures or background cues for trait prediction, we conducted a robustness evaluation by introducing synthetic noise into the test images (Figure 11). This experiment was designed to probe the sensitivity of each model to spatially localized perturbations and to reveal their respective dependence on biologically relevant versus contextual visual information. Given the limited size of the dataset, we applied a combination of additive and multiplicative noise types to simulate challenging visual conditions. The perturbations were introduced separately to the original and segmented test images under two distinct configurations:

- ResNet101: Received noisy versions of the raw images.
- CrossViT: Both branches received noisy versions of the raw images.
- AT-ViT: The small branch received noisy original images, while the large branch was fed noisy segmented images.

The noise configurations applied during robustness evaluation are detailed in Table 4. This table outlines the parameters used to simulate various perturbations, independently applied to background and plant regions at different intensity levels (0.9 for background and 0.75 for plant). For each of the five noise types (speckle, Gaussian, impulse, channel randomization/texture, and edge corruption) the corresponding distributions, scaling factors, and pixel corruption ratios are specified. The primary goal of this evaluation is to determine which model relies more heavily on plant structures versus background information for trait prediction. By selectively corrupting either the plant or the background, we can assess the degree to which each model focuses on bio logically meaningful regions. This aligns with the central objective of our work: to develop models that ground their predictions in plant morphology rather than in irrelevant contextual cues.

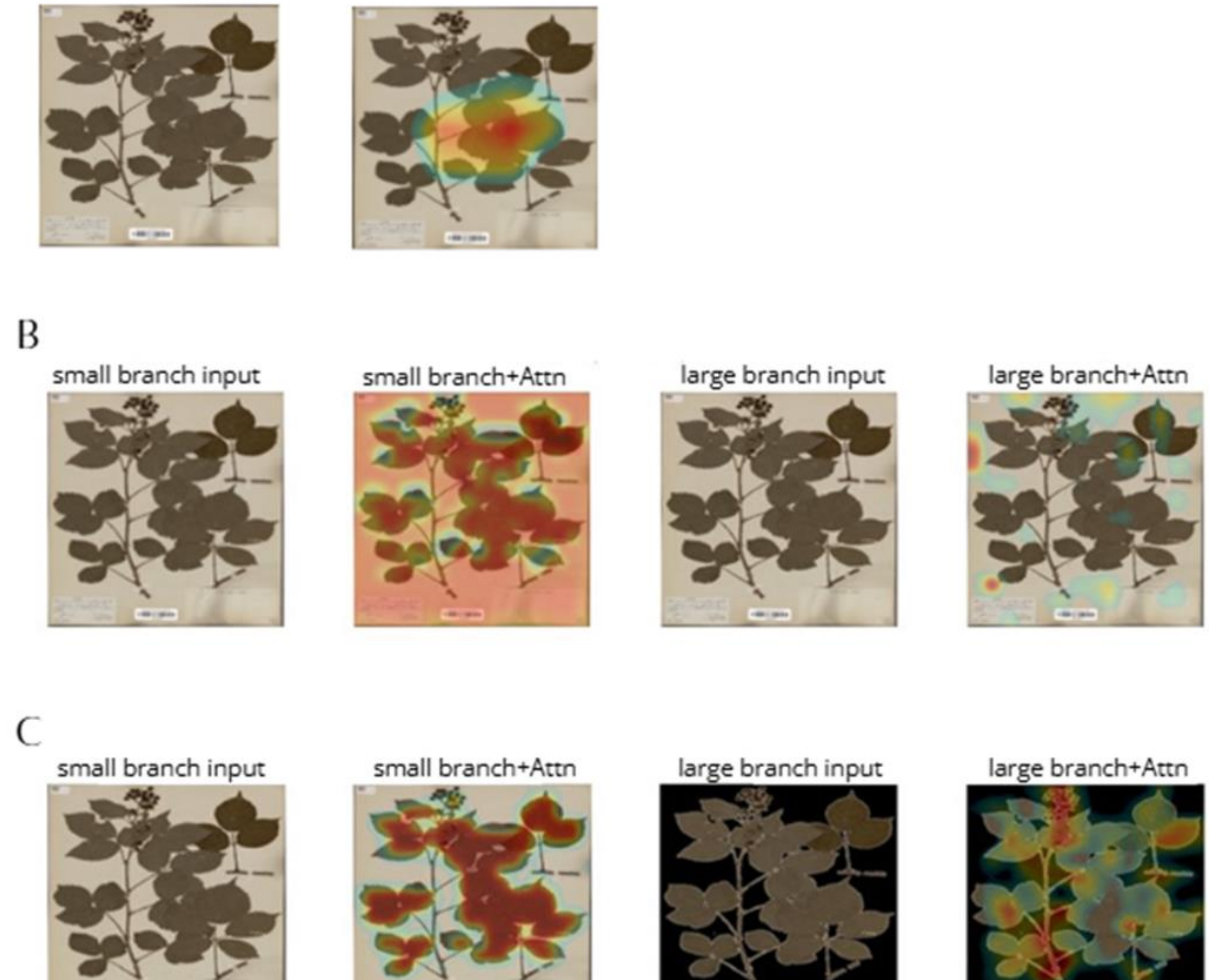

Figure 8: Visualization of activation and attention maps for the "leaves with an acute base" trait across ResNet101, CrossViT, and AT-ViT. **(A)** ResNet101 shows broad, low-resolution activations with correct prediction (True = 1.0 | Predicted = 1.0) **(B)** CrossViT exhibits diffuse attention that often focuses on background artifacts, resulting in an incorrect prediction (True = 1.0 | Predicted = 0.0). **(C)** AT-ViT consistently attends to informative leaf-base regions, yielding a correct prediction (True = 1.0 | Predicted = 1.0).

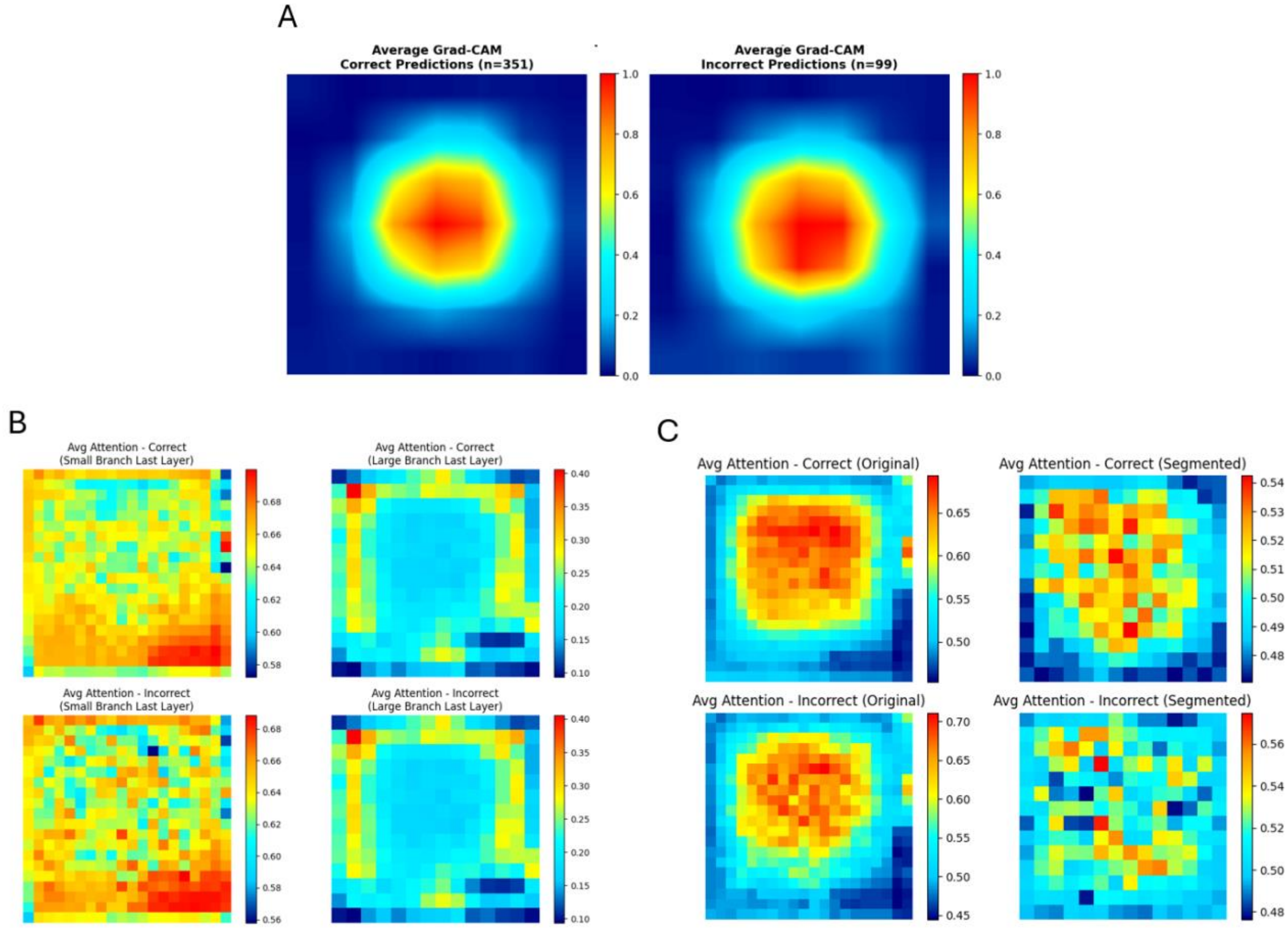

Figure 9: Average attention and saliency maps for the trait "leaves with an acute base."
**(A)** ResNet101**:** Grad-CAM highlights a small, centrally focused region for correct predictions, with incorrect predictions showing more diffuse activation. **(B)** CrossViT: Attention remains largely concentrated on peripheral image regions for both correct and incorrect predictions. **(C)** AT-ViT: Correct predictions exhibit centered, plant-relevant attention across both branches, while incorrect predictions display more dispersed patterns.

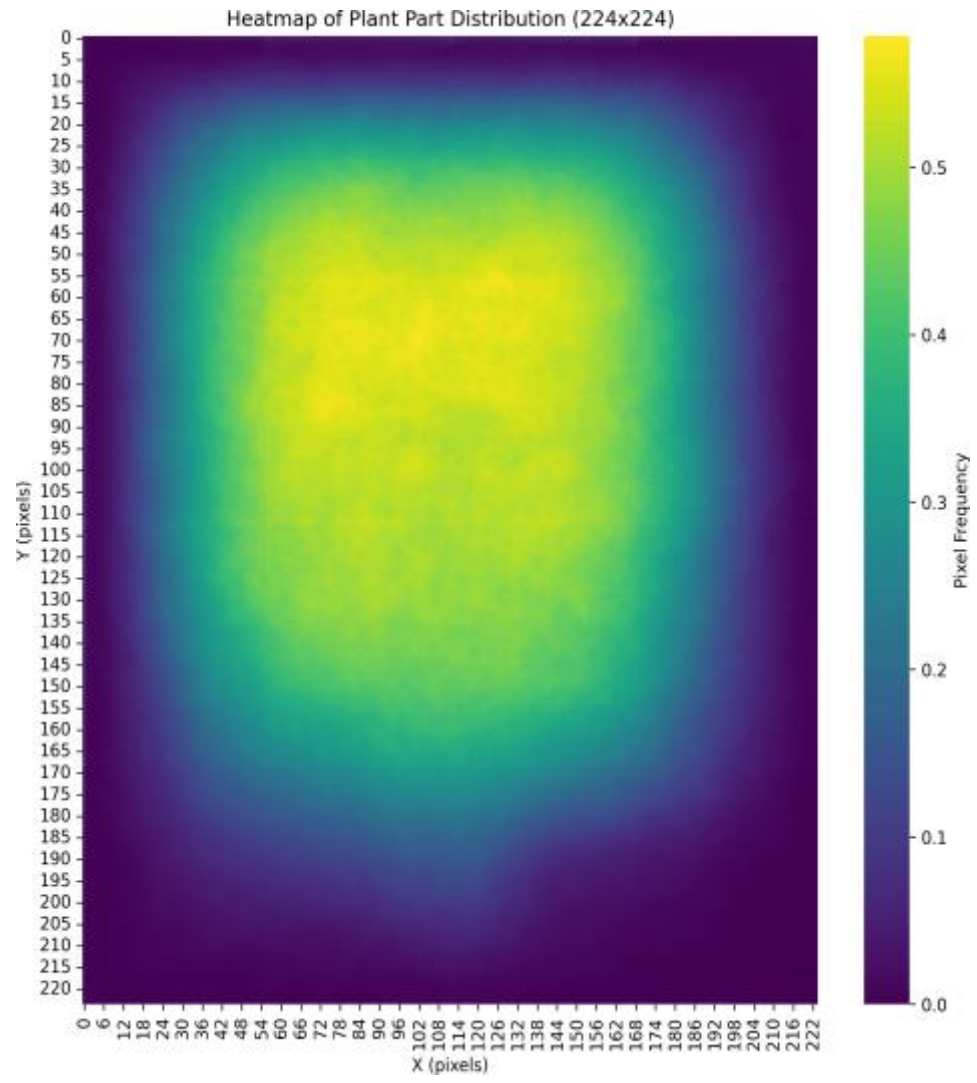

Figure 10: Pixel-wise heatmap of plant part distribution computed from all segmentation masks in the test dataset (resolution: 224×224). The visualization reveals a strong central concentration of plant pixels, indicating that herbarium specimens are typically centered within the image frame.

Table 4: Noise parameters used to simulate background and plant-region perturbations during robustness testing. Each noise type was applied separately to background and foreground (plant) regions with different intensity levels (0.9 for background, 0.75 for plant). The table specifies the noise distributions, pixel corruption percentages, and parameter scaling used to generate degradations across five types: speckle, Gaussian, impulse, channel randomization, and edge corruption.

| **Noise Type** | **Background Noise (Intensity: 0.9)** | **Plant Noise (Intensity: 0.75)** |
|---|---|---|
| Speckle | Gamma distribution, shape = 0.5, scale = 0.6 × 0.9 | Gamma distribution, shape = 1.2, scale = 0.3 × 0.75 |
| Gaussian | Mean = 0, std = 60 × 0.9 | Mean = 0, std = 35 × 0.75 |
| Impulse | 40% of pixels (0.4 × 0.9) replaced with values in [0, 255] | 20% of pixels (0.2 × 0.75) replaced with values in [0, 255] |
| Channel Randomization / Texture | 50% chance (0.5 × 0.9) to randomize 30% of one channel's pixels in [0, 255] | Texture noise: Gaussian, mean = 0, std = 20 × 0.75 |
| Edge Corruption | / | 30% of Canny edge pixels (0.3 × 0.75) randomized in [0, 255] |

We subsequently evaluated classification accuracy across the three trait prediction tasks under each noise configuration (Tables 5 and 6). The results indicate that when noise was applied predominantly to background regions, AT-ViT consistently outperformed both ResNet101 and CrossViT. This suggests that the patch-weighting mechanism in AT-ViT effectively suppresses irrelevant background information, thereby enhancing robustness to visual clutter. The advantage is particularly pronounced against ResNet101, with AT-ViT showing improvements of up to 32.32 percentage points under background noise conditions. In contrast, when noise was introduced directly within plant regions, the performance of AT-ViT degraded more noticeably than that of both baseline models. This behavior reflects the model's increased sensitivity to perturbations affecting biologically informative areas, an expected and desirable outcome that confirms AT-ViT's reliance on plant-specific visual cues for trait inference, rather than on spurious background correlations.

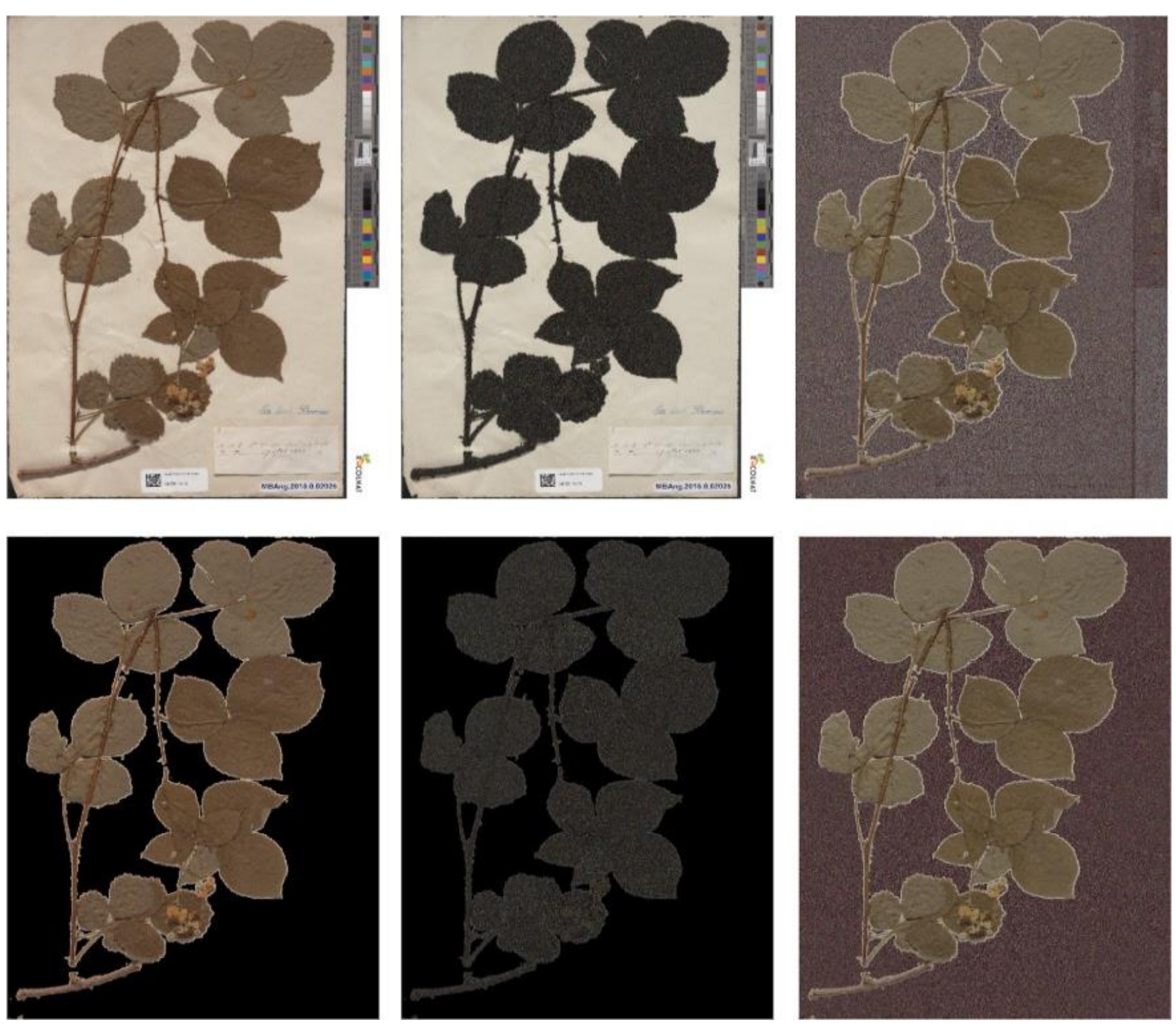

Figure 11: Top row: Original image, image with noise on plant, image with noise on background. Bottom row: Segmented image, segmented image with noise on plant, segmented image with noise on background.

Table 5: Impact of background and plant-region noise on classification accuracy for each trait, comparing the performance of ResNet101 and AT-ViT models under two perturbation scenarios: noise applied to background regions and noise applied to plant regions. The Δ columns indicate the accuracy difference between AT-ViT and ResNet101 under each noise condition.

| **Trait** | **Initial Accuracy** | | **ResNet101** | | **AT-ViT** | | **Δ** | |
|---|---|---|---|---|---|---|---|---|
| | *ResNet101* | *AT-ViT* | *Noise: Background* | *Noise: Plant* | *Noise: Background* | *Noise: Plant* | *Background* | *Plant* |
| Leaves with acuminate tips | 69.75 | 77.92 | 50.99 | 71.30 | 68.87 | 65.34 | +17.88 | −5.96 |
| Leaves with acute base | 77.55 | 79.33 | 73.56 | 72.67 | 75.11 | 70.44 | +1.55 | −2.23 |
| Thorns | 95.55 | 95.76 | 53.13 | 94.34 | 85.45 | 83.23 | +32.32 | −11.11 |

Table 6: Impact of background and plant-region noise on classification accuracy for each trait, comparing the performance of CrossViT and AT-ViT models under two perturbation scenarios: noise applied to background regions and noise applied to plant regions. The Δ columns indicate the accuracy difference between AT-ViT and CrossViT under each noise condition.

| **Trait** | **Initial Accuracy** | | **CrossViT** | | **AT-ViT** | | **Δ** | |
|---|---|---|---|---|---|---|---|---|
| | *CrossViT* | *AT-ViT* | *Noise: Background* | *Noise: Plant* | *Noise: Background* | *Noise: Plant* | *Background* | *Plant* |
| Leaves with acuminate tips | 77.04 | 77.92 | 63.80 | 75.94 | 68.87 | 65.34 | +5.07 | −10.60 |
| Leaves with acute base | 79.77 | 79.33 | 71.56 | 78.44 | 75.11 | 70.44 | +3.55 | −8.00 |
| Thorns | 96.36 | 95.76 | 83.03 | 93.33 | 85.45 | 83.23 | +2.42 | −10.10 |

## 6. Conclusion

In this study, we introduced AT-ViT, a dual-branch Vision Transformer architecture designed to integrate raw and segmented herbarium images via multi-scale, multi-view encoding and cross-attention fusion, to deliver segmentation-guided patch weighting. Experimental results demonstrate that AT-ViT consistently improves classification accuracy, enhances the alignment of attention with biologically meaningful regions, and exhibits increased robustness to background noise and visual clutter. Beyond quantitative performance, the interpretability of the model confirms its reliance on plant-specific structures rather than on irrelevant contextual artifacts, an essential property for developing reliable and explainable AI systems in the context of plant phenotyping.

Looking forward, we intend to further refine AT-ViT by incorporating spatial alignment constraints, such as Intersection-over-Union (IoU), directly into the training objective, thereby promoting closer correspondence between attention maps and true plant regions. We also plan to extend our evaluation to larger and more taxonomically diverse datasets, covering a wider spectrum of morphological traits. Beyond image-based classification, the segmentation-guided attention mechanisms introduced in AT-ViT could be directly leveraged within multimodal botanical question answering architectures, where grounded visual representations are crucial for reliable reasoning over herbarium data, as demonstrated in recent work on botanical visual–semantic modeling and question answering systems [28]. In addition, explicitly modeling and conveying information about non-plant elements, such as labels, color charts, and mounting artifacts could further improve the training of question answering models by enabling a clearer separation between botanical content and contextual metadata, following recent advances in contextual-aware learning for biodiversity data [29]. Ultimately, AT-ViT represents a step toward the development of scalable, interpretable, and biologically grounded deep learning frameworks capable of unlocking the full scientific potential of the world's herbarium collections.

**Acknowledgements**

This work was partially funded by the French National Research Agency (Agence Nationale de la Recherche; ANR) in the context of the e-Col+ project (ANR-21-ESRE-0053).

**Conflict of interest statement**

The authors declare that they have no known competing financial interests or personal relationships that could have appeared to influence the work reported in this paper.

**Funding information**

Agence Nationale de la Recherche (ANR), e-Col+ Project, Grant/Award Number: ANR-21-ESRE-0053

**Data availability statement**

The data that supports the findings of this study are available from the corresponding author upon request and will be released publicly upon acceptance.

**Code availability statement**

The source code used in this study is available from the corresponding author upon request and will be released publicly upon acceptance.

**Author Contributions**

Youcef Sklab led the conceptualization, data curation, formal analysis, and methodology development. He also managed the project, funding acquisition, provided resources, supervised the work, and contributed to validation, visualization, and both the original draft and subsequent revisions. Amani Sedrat contributed to methodology development, software implementation, and writing of the original draft. Takieddine Chehhat contributed to methodology development, software implementation, and writing of the original draft. Hanane Ariouat contributed to data curation and resource management. Abderrazak Sebaa contributed to writing of the original draft. Eric Chenin was involved in data curation, funding acquisition, project administration, and participated in manuscript revision. Edi Prifti contributed to funding acquisition and provided resources. Jean-Daniel Zucker contributed to the funding acquisition.